\pdfoutput=1
\documentclass[10pt,twocolumn,letterpaper]{article}

\usepackage{cvpr}
\usepackage{times}
\usepackage{epsfig}
\usepackage{graphicx}
\usepackage{amsmath}
\usepackage{amssymb}
\usepackage{multirow}

\usepackage[caption=false,font=footnotesize]{subfig}

\usepackage[pagebackref=true,breaklinks=true,letterpaper=true,colorlinks,bookmarks=false]{hyperref}

\cvprfinalcopy 

\def\cvprPaperID{12} 

\ifcvprfinal\fi
\begin{document}

\title{Is it Raining Outside? \\Detection of Rainfall using General-Purpose Surveillance Cameras}

\author{Joakim Bruslund\ Haurum, Chris H.\ Bahnsen, Thomas B. Moeslund\\
Visual Analysis of People Laboratory\\
Aalborg University, Denmark\\
{\tt\small \{joha, cb, tbm\}@create.aau.dk}
}

\maketitle
\thispagestyle{empty}

\begin{abstract}
In integrated surveillance systems based on visual cameras, the mitigation of adverse weather conditions is an active research topic. Within this field, rain removal algorithms have been developed that artificially remove rain streaks from images or video. In order to deploy such rain removal algorithms in a surveillance setting, one must detect if rain is present in the scene. 

In this paper, we design a system for the detection of rainfall by the use of surveillance cameras. We reimplement the former state-of-the-art method for rain detection and compare it against a modern CNN-based method by utilizing 3D convolutions. The two methods are evaluated on our new AAU Visual Rain Dataset (VIRADA) that consists of 215 hours of general-purpose surveillance video from two traffic crossings. The results show that the proposed 3D CNN outperforms the previous state-of-the-art method by a large margin on all metrics, for both of the traffic crossings. Finally, it is shown that the choice of region-of-interest has a large influence on performance when trying to generalize the investigated methods.

The AAU VIRADA dataset and our implementation of the two rain detection algorithms are publicly available at \url{https://bitbucket.org/aauvap/aau-virada} and \url{https://zenodo.org/record/4715681}.
\end{abstract}


\section{Introduction}
Varying weather and illumination conditions are a challenge for general-purpose outdoor surveillance systems \cite{buch2011review}. In order to deal with these challenges, several image and video optimization techniques have been proposed. The purpose of these techniques is to artificially remove haze and rain from the post-processed images or video. These techniques regularize the image in order to suppress the detrimental effects of the selected weather phenomena. As a side bonus, the appearance of objects of interest in the scene should be enhanced with respect to observation by a human observer or a computer vision system. 

\begin{figure}
    \centering
    \includegraphics[width=1.0\linewidth]{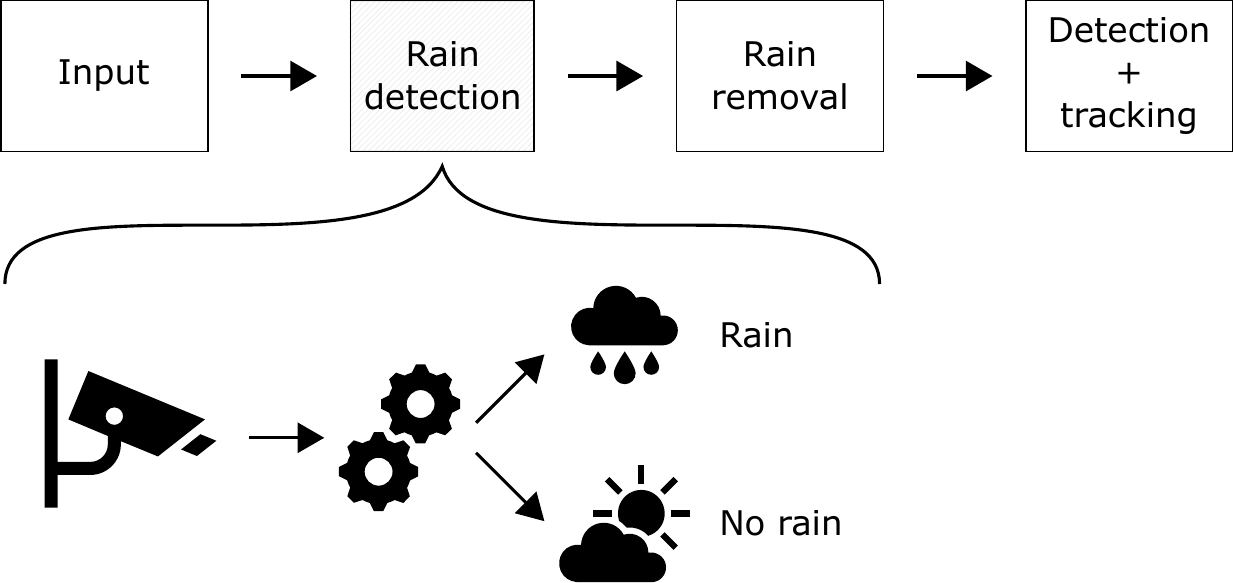}
    \caption{The proposed system at-a-glace. For rain removal algorithms to be effective in an integrated surveillance framework, the presence or absence of rain must be detected in a pre-processing step.}
    \label{fig:systemataglance}
\end{figure}

Because the haze and rain removal algorithms are built to improve the visibility of weather-beaten surveillance footage, they are reliant on other algorithms to detect the presence or absence of adverse weather conditions. 
As haze and rain removal algorithms are not general-purpose image enhancement algorithms, they will consistently deteriorate the output if either haze or rain streaks are not present in the scene. 
The decision process is especially important if real-time detection and tracking systems are built on top of the output of the rain removal algorithms \cite{bahnsen2018rain, li2019single}. An example of such a pipeline is illustrated in Figure \ref{fig:systemataglance}.

A prototype system for modelling the dynamic behaviour of outdoor surveillance scenes has been proposed in  \cite{alldieck2016context}. In this work, observations from a nearby weather station is used to guide a foreground detection algorithm.
However, the applicability of the method is limited by the dependence on external weather data. It is infeasible to place a weather station alongside every surveillance camera, and the correlation of the weather data and the observations by the camera is limited if the two sensors are not in close proximity.

We would ultimately want the cameras and algorithms to be as self-reliant as possible such that they may operate without external input. A prototype of such a system was built in \cite{hautiere2008meteorological} in which the input to a foreground segmentation algorithm was pre-processed by either a rain or fog removal algorithm.

The aim of this work is to investigate the use of existing surveillance cameras as surrogate rain detectors. As opposed to existing works on detection of rain, the primary purpose of our cameras is general-purpose surveillance. This implies that we have not adjusted the camera parameters to emphasize the appearance of rain drops. Our contributions are the following:
\begin{enumerate}
    \item A new publicly available rain dataset, the \textit{AAU Visual Rain Dataset} (VIRADA) \cite{virada}, consisting of 215 hours of recorded video from general-purpose surveillance cameras. Ground truth measurements of rainfall are provided by a nearby laser disdrometer and a conventional tipping-bucket rain gauge.
    \item A new rain detection algorithm based on the 3D CNN architecture of Tran \etal\cite{Tran2015}.
    \item An open-source implementation of the rain level detection algorithm of Bossu \etal\cite{Bossu2011}.
    \item Evaluation of the aforementioned methods on the proposed dataset. 
\end{enumerate}
\section{Related Work}
The detection of rain streaks in images and video has been tightly coupled to the removal of the very same rain streaks since Garg and Nayar published their studies on the appearance, detection, and removal of rain in the beginning of the millenium \cite{garg2003photometric, Garg2007}. The detection of rain streaks was seen as an intermediate step in order to suppress the streaks in the final, rain-removed image or video. 

Garg and Nayar noted that rain streaks appear brighter than their background and that the fast motion of the streaks imply that each streak is only visible in a single frame. Combined with the assumption of a quasi-static background, they detected rain streaks by using the photometric constraint:
\begin{align}
    \Delta I = I_n - I_{n-1} = I_n + I_{n+1} \geq c
    \label{eq:relw_photometric}
\end{align}
where $c$ is a threshold and $I_n$ denotes the image at frame $n$. Because the candidate streak image $\Delta I$ contains many false positives, different post-processing steps are required to filter the candidates. The photometric constraint of Equation \ref{eq:relw_photometric} is commonly used for the initial segmentation of rain streaks in many video-based rain removal algorithms \cite{Bossu2011, liu2009pixel, santhaseelan2015utilizing, tripathi2012video}.

Other approaches for detecting rain in the image space include morphological component analysis \cite{kang2012automatic} or matrix decomposition \cite{kim2015video}. These methods may be applied either on single images \cite{chen2017error,kang2012automatic} or video streams \cite{jiang2017novel, kim2015video, renvideo}.

Recently, the popularity of convolutional neural networks (CNNs) has reached the rain removal community. Amongst those works, some architectures explicitly produce a rain image that is used to refine the rain removal process \cite{liu2018erase, yang2017deep}. 
These methods might be applicable for stand-alone rain detection if their training process is tuned to the estimation of rain density and not the restoration of the rain-removed image.

A different approach for detecting rain in a video was proposed by Barnum \etal\cite{barnum2010analysis} who noted that the directional uniformity of the rain streaks was ideal for detection in the frequency space. They thus transferred the image into the Fourier domain where rain streaks were lying along an elongated ellipsis. However, the authors did not investigate if the volume of rain can be estimated in the frequency space.

For readers interested in a detailed overview of rain removal algorithms, we refer to a dedicated survey \cite{bahnsen2018rain}.

\subsection{Rain Density Estimation}
\label{subsec:rel_raindensity}
Bossu \etal\ \cite{Bossu2011} pioneered in using the detected rain streaks as a surrogate rain gauge. Motivated by the photometric constraint of Equation \ref{eq:relw_photometric}, they used a Mixture of Gaussians (MoG) \cite{stauffer2000learning} to model foreground and background objects. The candidate streaks were found by the following rule:
\begin{align}
    \Delta I = I_{FG} - I_{BG} \geq c    
    \label{eq:bossu_candidateextraction}
\end{align}
where $I_{FG}$ and $I_{BG}$ are the foreground and background images of the MoG model, respectively. False positives in $\Delta I$ are suppressed based on their size. The rotation of the remaining streaks are used to construct a Histogram of Orientation of Streaks (HOS). The appearance of the histogram is modelled using a Gaussian-uniform distribution, and the relationship between the Gaussian and the uniform parts of the distribution is used to detect the presence or absence of rain.

The work of Allamano \etal\cite{allamano2015toward} utilizes rigorous formulations of camera geometry to estimate the real-world volume of the detected rain drops. The photometric constraint is used to segment candidate streaks. The width and height of these streaks together with the focal length of the camera are used to estimate the rain rate.

The subsequent work of Dong \etal\cite{dong2017measurements} filters the candidate streaks by orientation and discards streaks not within the dominant orientation. Focused and unfocused streaks are distinguished based on intensity and edge information and used for estimating the length of each streak. The rain rate is estimated from the streak diameters by using a Gamma distribution.

Recent work of Jiang \etal\cite{jiang2019advancing} uses matrix decomposition to segment rain streaks from the background. The width of the detected streaks and the number of streaks are used to infer the rain rate. The authors use a Gamma distribution similar to \cite{dong2017measurements}.

The rain detection algorithm of Bossu \etal \cite{Bossu2011} is evaluated on footage from a general-purpose surveillance camera whereas the approaches of \cite{allamano2015toward,dong2017measurements,jiang2019advancing} are evaluated on footage from videos cameras whose parameters are tuned with the sole purpose of emphasizing the visual appearance of rain streaks.

\section{The AAU Visual Rain Dataset}\label{sec:dataset}

To the best of our knowledge, there is currently no publicly available dataset for benchmarking the detection of rain with general-purpose surveillance cameras. 
In order to fill this gap, we present the new publicly available \textit{AAU Visual Rain Dataset} (VIRADA) \cite{virada} that contains a total of 215 hours of surveillance video from two different locations in Aalborg, Denmark. The cameras are configured and positioned for traffic surveillance applications and not specifically configured for the task of detecting rain.

We obtain ground-truth precipitation data from two different rain gauges: a traditional, mechanical tipping-bucket rain gauge and a more advanced laser disdrometer \cite{loffler2000optical}. The two measurement devices are explained in the following.

\begin{figure}[!t]
    \centering
    \subfloat[Crossing1]{\includegraphics[width=0.65\columnwidth]{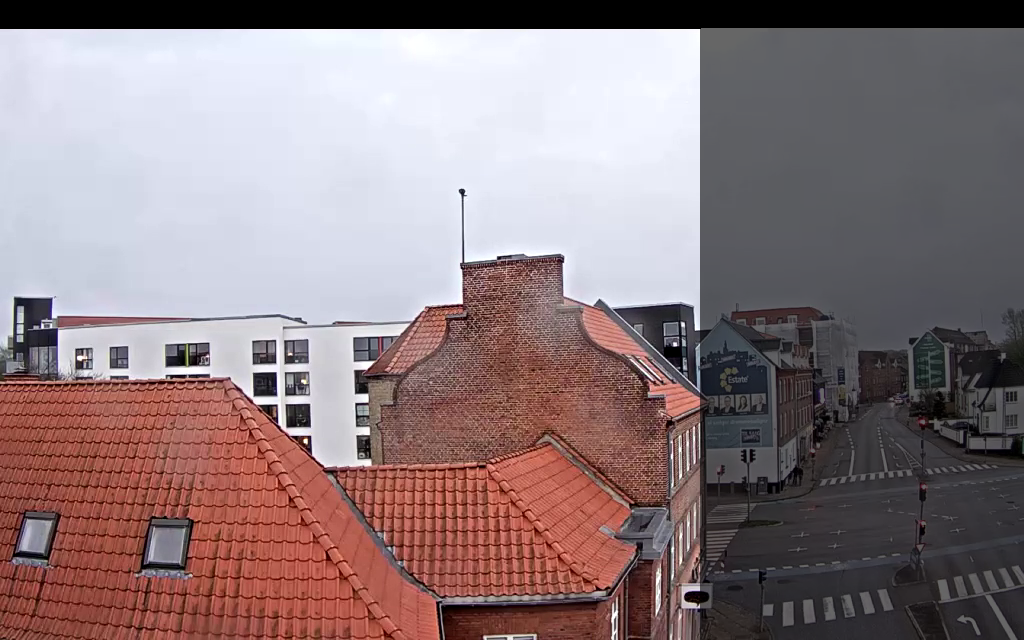}
    \label{fig:crossing1}}
    \hfill
    \subfloat[Crossing2]{\includegraphics[width=0.65\columnwidth]{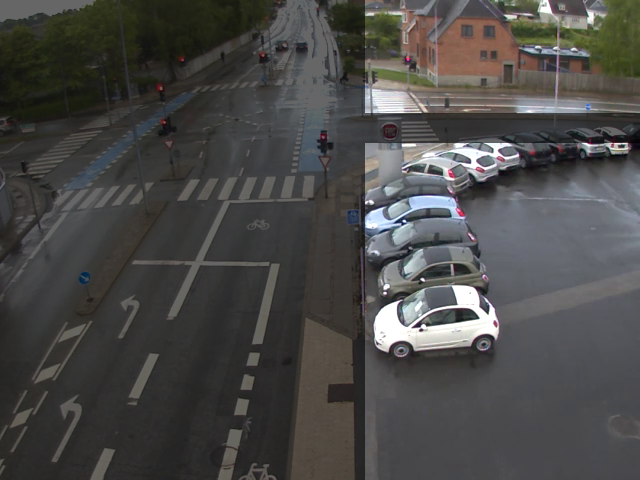}
    \label{fig:crossing2}}
    \caption{Sample views of the traffic crossings from the AAU VIRADA dataset. Discarded regions are shown with a semi-transparent overlay. We denote the upper region of Crossing2 as Crossing2-brick whereas the lower region is denoted as Crossing2-asphalt.}
    \label{fig:crossingsillustrated}
\end{figure}

\subsection{Rain Measurement Devices}
\paragraph{Tipping-bucket rain gauge}
In the tipping-bucket rain gauge, rain drops are collected by a funnel that channels the water into one of two seesaw-like buckets. When a bucket is full, it dumps the water and leaves the collection of water to the second bucket. An electric signal is generated whenever a container is full and the water is dumped. This type of measurement device is widely used and large networks of the devices have been utilized for different engineering domains \cite{MADSEN199829,MIKKELSEN19987}.

The resolution of the buckets is 0.2 mm which means that the bucket only tips once 0.2 mm of rain has passed trough the funnel. This implies that for low-intensity rainfall, \eg 0.1 - 2 mm/hour, it might take several minutes or even hours for the bucket to tip and for rain to be detected \cite{habib2001sampling}.
The signals generated by the tipping scales are post-processed in order to generate per-minute estimates of the precipitation level.

\paragraph{Laser disdrometer}
The laser disdrometer is an optical sensor that is capable of detecting single rain drops \cite{loffler2000optical}. A laser transmitter that transmits a sheet of light in free-air is located at the left side of the device. The sheet of light is detected on the right side of the device by an optical receiver. 
Because the laser disdrometer is capable of detecting individual rain drops, the temporal resolution is superior compared to the mechanical tipping-bucket rain gauge. Therefore, the laser disdrometer may be used for ground truth measurements when validating radar precipitation estimates \cite{AAUVandRadars, AAUVandNumerical}.

\subsection{Video Surveillance Sequences}
\begin{table*}[]
\centering
\begin{tabular}{lp{1.4cm}lp{0.8cm}p{1.7cm}p{1.6cm}lll}
\hline
Dataset             & Duration & Frames  & Frame & Native  & Cropped & Camera model & \multicolumn{2}{c}{Distance to gauge} \\ 
             & (hh:mm) &   & rate & resolution & resolution & & Mech.  & Laser \\ \hline 
Crossing1-trn & 87:38    & 9,276,654 & 30         & 1024 $\times$ 640        & 700 $\times$ 612 & AXIS Q1615-E & 580 m & 1230 m \\
Crossing1-val & 20:37    & 2,184,499 & 30         & 1024 $\times$ 640        & 700 $\times$ 612 & AXIS Q1615-E & 580 m & 1230 m \\
Crossing2-tst     & 106:59   & 9,463,287 & 25         & 640 $\times$ 480         & 276 $\times$ 338, 276 $\times$ 112 & AXIS M114-E & 1820 m & 970 m \\ \hline
\end{tabular}
\caption{Overview of the AAU VIRADA dataset. Mech.\ denotes the mechanical tipping-scale rain gauge whereas Laser denotes the laser disdrometer. The two noted cropped resolutions for the Crossing2-tst dataset are for the asphalt and brick crops, respectively.}
\label{tab:overviewstaticrain}
\end{table*}

\begin{table}[]
\centering
\begin{tabular}{lllll}
\hline
                    & \multicolumn{2}{c}{Detected rain \%} \\ \hline
Measurement device  & Laser             &  Mech.\ \\
                    \hline
Crossing1-trn & 19.67                     &17.97                \\
Crossing1-val & 20.86                     & 14.63                          \\
Crossing2-tst     & 8.68                          & 1.65                                    \\ \hline
\end{tabular}
\caption{Overview of the ratio of detected rain for the AAU VIRADA dataset, per measurement device. Mech. denotes the mechanical tipping-scale rain gauge whereas Laser denotes the laser disdrometer.}
\label{tab:overviewlabelsplit}
\end{table}

We have collected video footage from two different traffic crossings, in the following denoted as Crossing1 and Crossing2. The Crossing2 sequence is recorded in 2013 using an AXIS M114-E camera whereas the Crossing1 sequence is recorded in 2018 using a newer AXIS Q1615-E camera. Sample footage from the two crossings are shown in Figure \ref{fig:crossingsillustrated}. In order to ease the task at hand, we only consider regions in the video with few moving objects due to the following reasons:

\begin{enumerate}
    \item The detection of rain from general-purpose surveillance cameras is hard. In order to solve the problem, we should first solve the simpler sub-problem.
    \item For many surveillance applications, it is possible to select a region of interest where objects are mostly static, for instance the facade of a building.
    \item The selection of a region with no moving road users allows the public release of the dataset.
\end{enumerate}

The discarded regions are masked out in Figure \ref{fig:crossingsillustrated} with a semi-transparent overlay. A single region is chosen for Crossing1, whereas Crossing2 is split into two regions, Crossing2-brick and Crossing2-asphalt. Due to privacy concerns, only the parts of Crossing2-asphalt with no pedestrians are publicly available. 

The footage from Crossing1 is split into a training (trn) and a validation (val) set whereas the Crossing2 is used in its entirety for testing. An overview of the dataset and the distance from the cameras to the dedicated rain measurement devices is found in Table \ref{tab:overviewstaticrain}. The ratio of rain detected by the rain measurement devices is listed in Table \ref{tab:overviewlabelsplit}.

\section{Methods}
We evaluate two different methods for detecting rain from surveillance video: Bossu \etal rain detection \cite{Bossu2011} and the C3D CNN architecture by Tran \etal \cite{Tran2015}, each representing a different paradigm. The rain-detection algorithm by Bossu \etal represents a hand-crafted algorithm specifically designed for this task. On the other hand, the C3D CNN was originally developed for action recognition, scene classification and object recognition where temporal information is encoded through 3D network layers.

\subsection{Bossu Rain Detection}
\begin{figure}
    \centering
    \includegraphics[width=\linewidth]{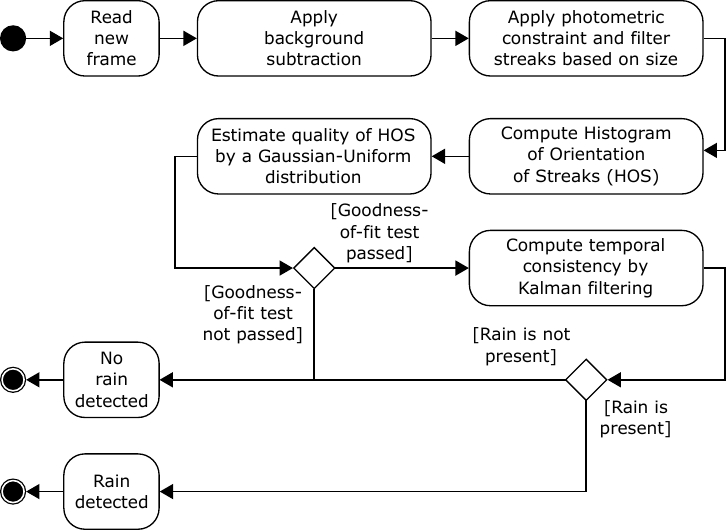}
    \caption{Activity diagram of the rain detection algorithm by Bossu \emph{et al.} \cite{Bossu2011}.}
    \label{fig:bossu_activity}
\end{figure}

Bossu \etal \cite{Bossu2011} propose a rain detection algorithm which in its core is based on detecting the approximate angle of the rain streaks in an image. This is done by assuming the rain streaks to be Gaussian distributed around a center angle $\theta$ with uncertainty $d\theta$. Based on the estimated distribution parameters, it can be decided whether rain is present or not. The algorithm is illustrated in Figure~\ref{fig:bossu_activity}. 

Background subtraction and computation of candidate streaks are described in Section \ref{subsec:rel_raindensity} and Equation \ref{eq:bossu_candidateextraction}. In the following, the remaining steps of the algorithm are described. We refer the reader to the original article \cite{Bossu2011} for a complete reference.

\paragraph{Histogram of Orientation of Streaks}
In order to determine the general orientation of the rain streaks, we create a 180-bin histogram in the range [0, 179]. We approximate the rain streak BLOBs as ellipses. To determine the orientation of the ellipses, we compute the geometric moments of the $i$th BLOB based on the central second-order moments, $m_i^{20}$, $m_i^{11}$, and, $m_i^{02}$. This leads to the calculation of $\theta_i$, $d\theta_i$, and, $w_i$ as shown in Equation~\ref{eq:theta}-\ref{eq:weight}. $dm$ is an empirically chosen scaling constant of the uncertainty, and $\lambda_i^1$ is the largest eigenvalue of the matrix 
$\begin{bmatrix}\label{eq:momMat}
m_i^{20} & m_i^{11} \\
m_i^{11} & m_i^{02}
\end{bmatrix}$.

\begin{equation}\label{eq:theta}
\theta_i = \frac{1}{2}\tan^{-1} (\frac{2m_i^{11}}{m^{02}_i-m^{20}_i})
\end{equation}

\begin{equation}\label{eq:dtheta}
d\theta_i = \frac{\sqrt[]{(m_i^{02} - m_i^{20})^2 + 2(m_i^{11})^2}}{(m_i^{02} - m_i^{20})^2 + 4(m_i^{11})^2}dm
\end{equation}

\begin{equation}\label{eq:weight}
w_i = \sqrt[]{\lambda_i^1}
\end{equation}

The Histogram of Orientation of Streaks (HOS) can then be computed, where $B$ is the total amount of BLOBs:\\
\begin{equation}\label{eq:hos2}
h(\theta) = \sum^{B}_{i}\frac{w_i}{d\theta_i\sqrt[]{2\pi}}e^{-\frac{1}{2}(\frac{\theta-\theta_i}{d\theta_i})^2}
\end{equation}

\begin{figure*}[!ht]
    \centering
    \includegraphics[width=1.0\textwidth]{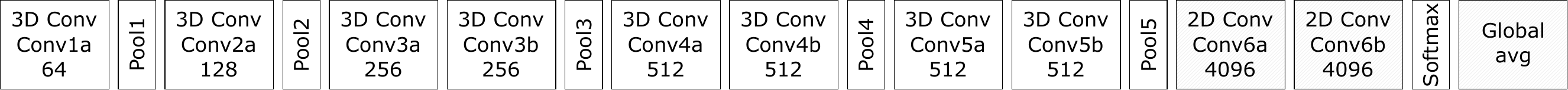}
    \caption{Overview of the modified C3D CNN architecture. 3D conv and 2D conv denotes 3D and conventional 2D convolutions, respectively. Pool denotes max pooling and the number of filters are denoted in the bottom of each box. Our modifications of the network are marked in grey. Figure adapted from \cite{Tran2015}.}
    \label{fig:C3D-architecture}
\end{figure*}
\paragraph{HOS quality estimation}
The HOS is built on the assumption that all BLOBs in the image are representations of actual rain streaks and that the orientation of the rain streaks follows a Gaussian distributed. BLOBs that represent noise in the foreground segmentation or stem from other non-rain scene elements thus have to be removed.
We assume that non-rain BLOBs in the image are uniformly distributed which means that the HOS can be modeled by a Gaussian-uniform distribution:
\begin{equation}\label{eq:hosGU}
y(\theta) \sim \Pi \ \mathcal{N}(\theta|\mu,\sigma) + (1-\Pi)\mathcal{U}_{[0,179]}(\theta)
\end{equation}
where $\Pi$ denotes the ratio of the Gaussian distribution in the HOS. We assume that the rain streaks contribute to the Gaussian part of the distribution.

The parameters $\mu$, $\sigma$ and $\Pi$ are estimated through an Expectation-Maximization (EM) algorithm based on the computed HOS in Equation~\ref{eq:hos2}.

When the EM algorithm has converged, the final HOS has to be evaluated. If the EM determined parameters does not result in a distribution that is close to the actual observed histogram $h(\theta)$, then the frame is discarded from further processing. In order to quantify this statement, a Kolomogrov-Smirnov goodness-of-fit test is performed:
\begin{equation}\label{eq:ks}
D = \sup_\theta |F_n(\theta) - F(\theta)|
\end{equation}
where $F({\theta})$ is the cumulative distribution function of a Gaussian with the EM-estimated parameters and $F_n(\theta)$ is the accumulated HOS histogram $h(\theta)$, an empirical cumulative distribution function.

If $D$ is above some threshold $D_c$, it is determined that it is not raining in the frame. If $D \leq D_c$, we estimate the temporal consistency in the following.

\paragraph{Temporal consistency}
In order to be temporally consistent in the detection of rain, a Kalman filter is used to track and smooth the EM estimated parameters. If the estimated Gaussian ratio, $\Pi$, is larger than some threshold $\Pi_\text{rain}$, the Kalman filter is updated and rain is detected.

\subsection{C3D Convolutional Neural Network}
The use of 3D convolutions to encode temporal information has been an ongoing topic in the research community since Tran \etal \cite{Tran2015} proposed their C3D architecture. Some of the recent advances include more complex networks \cite{NonLocal2018}, residual networks \cite{Hara2018}, and separable convolutional layers \cite{Tran2018}. However, in order to provide a baseline for video-based rain detection, we investigate the well-established original C3D network.

The C3D CNN architecture builds on the concept of using series of consecutive video frames as input and  utilizing 3D convolutions instead of 2D convolutions. Specifically, each input is changed from a 3D tensor of size $c\times h\times w$ to a 4D tensor of size $c\times l\times h\times w$. The parameters $c$, $w$, and $h$ are the number of channels and the height and width of the the input whereas $l$ is the length of the input sequence. The receptive field of the filters is also changed from $k\times k$ to $d\times k\times k$, where $k$ and $d$ are the spatial and temporal extent of the filter, respectively.

The original C3D network ends with two fully-connected layers of size 4096, a dropout rate of 50\%, and a softmax layer. This has the disadvantage of forcing a specific input size for the image, meaning the input should be cropped or resized. In order to get a single output for the entire image, we convert the network to a fully-convolutional network (FCN) by replacing the fully-connected layers with 2D convolutional layers and adding a global averaging layer on top of the softmax layer. The two new convolutional layers, Conv6a and Conv6b, have filter sizes of $512\times 4\times 4$ and $4096\times 1\times 1$ in order to function in a similar way as a fully-connected layer. The modified network architecture is shown in Figure~\ref{fig:C3D-architecture}. 
By converting the network to a FCN, it effectively applies a sliding window approach with a spatial stride of 32.

\section{Implementation}
In the following, we will guide the reader through the most important implementation details of the investigated methods. Due to the inherently higher precision of the laser disdrometer, as discussed in Section~\ref{sec:dataset}, only labels from the laser disdrometer are utilized for training and evaluation in this work.  Our implementation is publicly available.

\subsection{Bossu}
As the method from Bossu \etal \cite{Bossu2011} was not publicly available, we have implemented it from scratch in C++ with the OpenCV framework. We use the first 500 frames of each video to initialize the background model of MoG. The EM algorithm is initialized according to the instructions of Bossu \etal \cite{bossu2009}. The process and measurement noise covariance matrices of the Kalman filter are initialized with a variance of 0.01 and 0.1, respectively, and a covariance of 0, as per the original authors \cite{Bossu2011}.\\

In order to determine the remaining parameter values, we perform a parameter search on six video snippets from the Crossing1 dataset with an equal amount of rain and non-rain videos. The duration of the snippets vary from 5 to 20 minutes. A total of 9600 parameter combinations are investigated and the specific values included in the search are shown in Table~\ref{tab:BossuParamSearch}.
We selected the final parameters based on the following criteria:

\begin{itemize}
    \item For rain sequences, the Bossu algorithm should detect rain for at least 60 \% of the frames, preferably more.
    \item For no-rain sequences, the Bossu algorithm should detect rain for maximum 40 \% of the frames, preferably less.
\end{itemize}

The collection of parameters that performs most consistently under these criteria is listed in the rightmost column of Table~\ref{tab:BossuParamSearch}.

\begin{table}
\begin{center}
\begin{tabular}{llp{1.5cm}}
\hline
Parameter & Search space & Selected value\\
\hline
MoG warm-up frames & [500] & 500\\
$c$ & [3, 5] & 3\\
Minimum BLOB size & [4] & 4\\
Maximum BLOB size & [50:50:200] & 200\\
$dm$ & [0.5:0.5:2.0] & 0.50\\
EM max iterations & 100 & 100\\
$D_c$ & [0.01:0.01:0.20] & 0.19\\
$\Pi_\text{rain}$ & [0.20:0.02:0.50] & 0.40\\
\hline
\end{tabular}
\end{center}
\caption{Values and search space for the Bossu parameter search. The ranges in the search space follow the python convention, with [3,5] being a list of parameters and  [0.5:0.5:2.0] referring to values in the range from 0.5 to 2.0 with an interval of 0.5}
\label{tab:BossuParamSearch}
\end{table}

\subsection{C3D}
We train the C3D network from scratch on the training/validation split of the Crossing1 videos, utilizing 2 Tesla V100 graphic cards. The network is trained as a binary classification problem on $112\times 112$ sized crops from the videos in order to maintain a reasonable batch size. We load video sequences with a temporal stride of 8 frames. 

We use a stochastic gradient descent optimizer with momentum, weight decay, and a step-based learning rate scheduler for training. The learning rate scheduler multiplies the learning rate by $\gamma$ every $s$ epochs. The set of hyperparameters used during training are listed in Table~\ref{tab:C3DHyper}.
The PyTorch deep learning framework \cite{Pytorch} is used for creating and training the model and the sequence loading was handled using the NVIDIA Video Loader framework \cite{nvvl}.
We augment the data by randomly chosen crops and random flipping along the vertical axis with 50\% chance.

\paragraph{Validation}
For the Crossing1 videos, the FCN structure results in a 19 $\times$ 17 patch grid being investigated while for the Crossing2-tst videos, a 6 $\times$ 8 and 6 $\times$ 1 patch grid is investigated for the asphalt and brick crops, respectively. Each patch outputs a vector containing the output of the softmax layer. These vectors are subsequently averaged and thresholded in order to get the final binary prediction for the video sequence.

\begin{figure}[!ht]
\centering
\includegraphics[width=\linewidth]{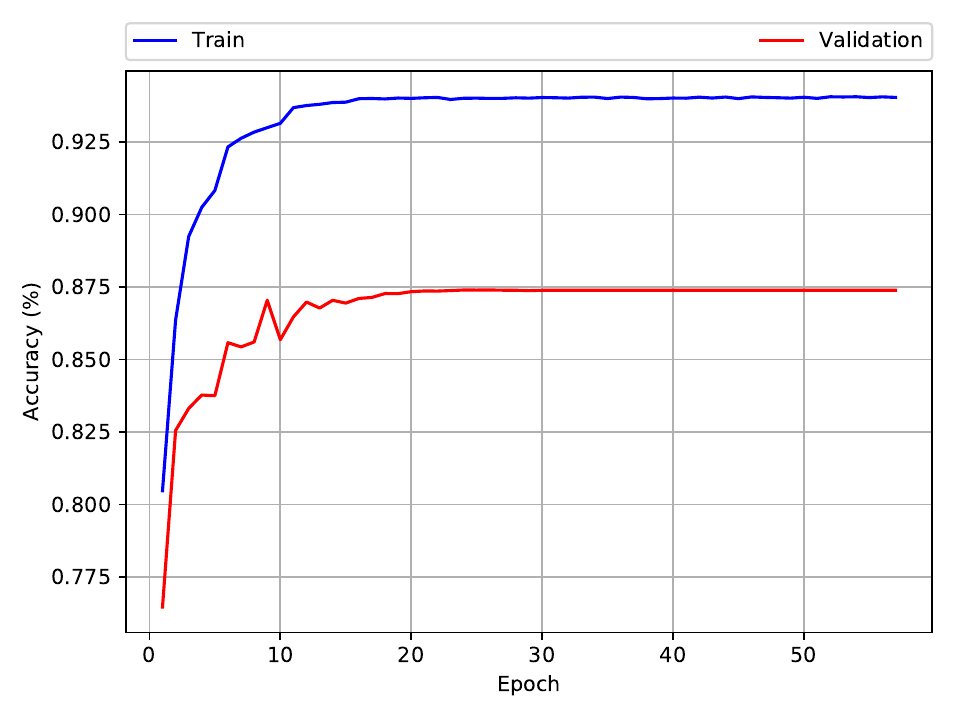}
\caption{Average accuracy per epoch for the trained C3D CNN.}
\label{fig:C3Dacc}
\end{figure}

\begin{figure}[!ht]
\centering
\includegraphics[width=\linewidth]{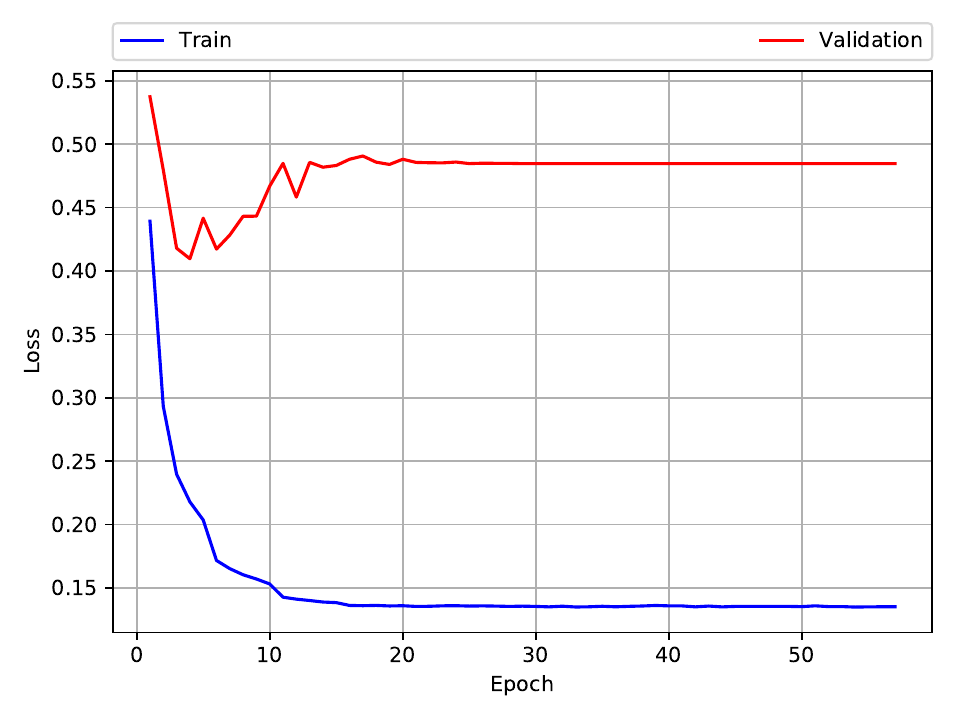}
\caption{Average loss per epoch for the trained C3D CNN.}
\label{fig:C3Dloss}
\end{figure}

The network is trained for 57 epochs and reached a training accuracy of $94.03\%$ and a validation accuracy of $87.38\%$. The accuracy and loss plots are shown in Figure~\ref{fig:C3Dacc} and Figure~\ref{fig:C3Dloss}.

\begin{table}
\begin{center}
\begin{tabular}{ll}
\hline
Hyperparameter & Value \\
\hline
Batch size & 128 \\
Sequence stride & 8\\
Learning rate & 0.01 \\
Momentum & 0.9 \\
Weight decay & 0.0001\\
$\gamma$ & 0.1\\
$s$ & 5\\
\hline
\end{tabular}
\end{center}
\caption{C3D hyperparameters.}
\label{tab:C3DHyper}
\end{table}

\section{Experimental Results}

\begin{table*}[!ht]
\centering
\begin{tabular}{llccccccc}
\hline
Sequence &Method       & TP     & TN     & FP     & FN     & Acc    & F1     & MCC    \\ \hline 
\multirow{3}{*}{Crossing1-trn}&C3D-FCN          &  215244 & 923596 &   7766 &  12802 &\textbf{0.9823}  & \textbf{0.9544}  & \textbf{0.9435}  \\
&C3D-Center          &   202932 &  920692 &  10672 &  25114  & 0.9691 & 0.9190 & 0.9007 \\
&Bossu-EM     &  1096369 & 3498967 & 3915541 &  719777  & 0.4978  & 0.3211  & 0.0603 \\
&Bossu-Kalman &  1119361 & 3460093 & 3954415 &  696785  & 0.4961 & 0.3249 & 0.0663  \\ \hline
\multirow{3}{*}{Crossing1-val}&C3D-FCN          &  30126 & 208447 &   7405 &  27042  & \textbf{0.8738}  & \textbf{0.6362} & \textbf{0.5821}  \\
&C3D-Center          &   25320 &  199555 &  16298 &  31847  & 0.8237 & 0.5126 & 0.4159 \\
&Bossu-EM     & 263008 & 912983 & 805113 & 192395  & 0.5411  & 0.3453 & 0.0887 \\
&Bossu-Kalman & 267253 & 909237 & 808859 & 188150  & 0.5413  & 0.3490 & 0.0945 \\\hline
\multirow{3}{*}{Crossing2-asphalt}&C3D-FCN          &      0 & 1069231 &       0 & 102717 & \textbf{0.9124} & 0.0000 & 0.0000  \\
&C3D-Center          &   245 &  1039578 &  40409 &  102474  & 0.8792 & 0.0034 &  -0.0837 \\
&Bossu-EM     & 224853 & 6335804 & 2257136 & 591994 & 0.6972 & 0.1363 & \textbf{0.0080}  \\
&Bossu-Kalman & 234181 & 6264711 & 2328229 & 582666 & 0.6907 & \textbf{0.1386} & 0.0010  \\ \hline
\multirow{3}{*}{Crossing2-brick}&C3D-FCN          &   72619 &  729561 &  350381 &  30095  & \textbf{0.6783} & 0.2763 &  0.2248 \\
&C3D-Center          &   75690 &  720369 &  359557 &  27024  & 0.6731 & \textbf{0.2814} &  \textbf{0.2359} \\
&Bossu-EM     &  281084 & 5837499 & 2755441 & 535763  & 0.6502 & 0.1459  & 0.0141  \\
&Bossu-Kalman &  290583 & 5762519 & 2830421 & 526264  & 0.6433 & 0.1476  & 0.0158  \\ \hline
\end{tabular}
\caption{Rain detection results on the AAU VIRADA dataset, using labels from the laser disdrometer.}
\label{tab:results-laser}
\end{table*}

The trained algorithms are evaluated on the AAU VIRADA dataset presented in Section~\ref{sec:dataset}. In order to get a quantitative measure of the rain detection performance, several metrics are used. First, the values of the confusion matrix are reported: True Positive (TP), True Negative (TN), False Positive (FP), and False Negative (FN). In this case a True Positive is when rain is correctly detected, while a True Negative is when no rain is correctly detected. Based on these quantities, the \textbf{Accuracy} (Acc), \textbf{F1 Score} (F1), and \textbf{Matthews Correlation Coefficient} (MCC) are calculated as follows:

\begin{align}
    \text{Acc} &= \frac{\text{TP} + \text{TN}}{\text{TP}+\text{TN}+\text{FP}+\text{FN}}     \label{eq:acc} \\
    \text{F1} &= \frac{2\ \text{TP}}{2\ \text{TP} + \text{FP} + \text{FN}}     \label{eq:F1} \\
    \text{MCC} &= \frac{\text{TP}\cdot \text{TN} - \text{FP}\cdot \text{FN}}{\sqrt{(\text{TP}+\text{FP})(\text{TP}+\text{FN})(\text{TN}+\text{FP})(\text{TN}+\text{FN})}}
    \label{eq:MCC}
\end{align}\\

The accuracy metric is an often used metric which directly indicates the correct percentage of assigned frames. However, it is not a good indicator for imbalanced datasets and can in these cases be misleading. The F1 score and MCC try to counteract this problem. 

The F1 score provides a metric where true negatives are not considered, which for datasets skewed towards a high amount of trivial true negatives results in a more fair representation.

In the same sense, the MCC metric tries to provide a fair single value representation of the confusion matrix, even for imbalanced datasets, by providing a value in the range [-1; 1]. -1 indicates total disagreement, 0 indicates pure guesswork, and 1 indicates perfect predictions. If any of the sums in the denominator results in 0, the resulting value is set to 0. The MCC metric is a measure which have been recommended for computational biology and biomedical and health informatics due to its built-in considerations for both positive and negative predictions in imbalanced datasets \cite{Chicco2017}. MCC will be used as the primary evaluation metric.

The method evaluation is not a one-to-one comparison, as the Bossu rain detection algorithm works on a per-frame basis while the C3D CNN analyse 16 frames at a time, with a 8 frame stride. Therefore, there will be fewer predictions for the C3D CNN. In order to demonstrate the effect of the FCN structure, the C3D CNN will be evaluated when applied on the entire frame, utilizing the FCN structure, and evaluated with a $112\times112$ center patch of the frame. These are denoted \textit{C3D-FCN} and \textit{C3D-Center}, respectively. Furthermore, the Bossu algorithm is evaluated in two ways, by investigating the rain detection capabilities when using either the per-frame EM estimated HOS parameters, or the Kalman smoothed HOS parameters. These are denoted \textit{Bossu-EM} and \textit{Bossu-Kalman}, respectively. The laser disdrometer labels have been converted from per-minute to per-frame representations. The results are shown in Table~\ref{tab:results-laser}. The best performing metrics are highlighted in bold.

From the results it is evident that the C3D CNN outperforms the Bossu algorithm on all of the Crossing1 videos. 
The Bossu rain detector algorithm provides nothing more than a random guess, as shown by the accuracy values near 50\% and MCC values near 0. On the other hand, the modified C3D CNN achieves a near perfect accuracy of 98 \% and MCC of 0.94 on the training set whereas a  87\% accuracy and MCC of 0.58 is scored on the validation set. This indicates that while it performs well, performance can be improved, as shown by the large amount of false negative predictions. Comparatively, if only the center is evaluated with the C3D CNN, the performance drops drastically to a MCC of 0.90 and 0.41 on the training and validation sets, respectively. 
As we trained the C3D CNN on a subset of the Crossing1 dataset and determined the parameters of the Bossu algorithm on the very same dataset, the difference in performance is striking.

On the Crossing2 videos, two crops are tested: One with asphalt background and one with a brick house background. It is shown that neither the C3D CNN nor the Bossu algorithm generalize well when tested on the asphalt background. The C3D CNN evaluating the entire frame predicts no rain, while the Bossu algorithm predicts rain approximately one third of the time. The C3D CNN evaluated on just the center patch does predict rain in some instances, but due to the large discrepancy between true and false predictions, it results in a MCC of -0.08. When tested on the brick house background, however, the C3D CNN outperforms the Bossu rain detector on all metrics. This indicates that the C3D CNN can generalize somewhat when evaluated on surfaces similar to the one it was trained on. It is also found that by just evaluating the center patch with the C3D CNN, the MCC increases by 0.01.

The results also show that the Bossu algorithm works better on the brick house background but that the performance is still affected by a large amount of false positives and negatives. 

We hypothesize that the reason C3D-Center performs better than C3D-FCN on the Crossing2 data, is due to the dynamic effects that occurs in the regions. In Crossing2-asphalt there are many cars with reflections, while in Crossing2-Brick there are pedestrians walking by along the sidewalk. By using just the center patch, some of these dynamic effects may be avoided. Further investigation is needed in order to be certain.\\
\section{Conclusion}
In this work we investigated detection algorithms for general-purpose surveillance cameras. The current state-of-the-art method and a data-driven 3D CNN method were implemented and compared on a new publicly available dataset, the AAU Visual Rain Dataset (VIRADA), consisting of 215 hours from two separate traffic crossings, making it by far the biggest rain dataset captured by general purpose surveillance cameras. A subset of one of the traffic crossing videos was used to train the algorithms.  When testing on unseen data from the traffic crossing the algorithms were trained on, we found that our modified 3D CNN algorithm outperformed the previous state-of-the-art method. However, when testing on data from a new traffic crossing, the performance of the algorithms were dependent on the similarity of the investigated region-of-interest and the training data. Using a similarly textured region-of-interest, our 3D CNN outperformed the previous state-of-the-art by a large margin. Comparatively, when using a region-of-interest with a very different kind of texture, our 3D CNN failed to function. From these observations it is clear that our modified 3D CNN outperforms the previous state-of-the-art, but also that the task of rain detection for general-purpose surveillance cameras is not yet solved.

Future work could include an in-depth comparison between the laser disdrometer and the mechanical tipping-scale rain gauge in order to determine the effect of the label quality on the evaluated results.

{\small
\bibliographystyle{ieee_fullname}
\bibliography{rainDetection}
}

\end{document}